\documentclass[conference]{IEEEtran}
\IEEEoverridecommandlockouts
\usepackage{cite}
\usepackage{amsmath,amssymb,amsfonts}
\usepackage{algorithmic}
\usepackage{graphicx}
\usepackage{textcomp}
\usepackage{multirow}
\usepackage[table,xcdraw]{xcolor}
\def\BibTeX{{\rm B\kern-.05em{\sc i\kern-.025em b}\kern-.08em
    T\kern-.1667em\lower.7ex\hbox{E}\kern-.125emX}}

\DeclareMathAlphabet\mathbfcal{OMS}{cmsy}{b}{n}

\begin{document}

\def \Hm {\mathbf{H}_{m}}
\def \Hh {\mathbf{H}_{h}}
\def \Ym {\mathbf{Y}_{m}}
\def \Yh {\mathbf{Y}_{h}}
\def \c  {\frac{1}{2}}
\def \crho {\frac{\rho}{2}}
\newcommand{\vseq}{\\[4pt]}
\newcommand{\minp}[1]{\underset{#1}{\text{min}}} 
\newcommand{\mintwo}[2]{\underset{#1,#2}{\text{min}}} 
\newcommand{\normF}[1]{\lVert #1 \rVert_F}
\newcommand{\normtwo}[1]{\lVert #1 \rVert_2^2}
\newcommand{\normone}[1]{\lVert #1 \rVert_{1}}
\newcommand{\normzero}[1]{\lVert #1 \rVert_{0}}
\newcommand{\normonet}[1]{\lVert #1 \rVert_{1}}
\newcommand{\mtx}[1]{\mathbf{#1}}
\newcommand{\mtxs}[2]{\mathbf{#1}_#2}
\newcommand{\rvect}[1]{\begin{bmatrix} #1 \end{bmatrix}}
\newcommand{\edwin}[1]{\textcolor{red}{#1}}
\newcommand{\ideas}[1]{\textcolor{blue}{#1}}
\newcommand{\emmanuel}[1]{\textcolor{orange}{#1}}
\newcommand{\emmaideas}[1]{\textcolor{purple}{#1}}
\newcommand{\remove}[1]{\textcolor{cyan}{#1}}

\title{Fast Disparity Estimation from a Single Compressed Light Field Measurement 

}

\author{\IEEEauthorblockN{1\textsuperscript{st} Emmanuel Martínez}
\IEEEauthorblockA{\textit{Department of Computer Science} \\
\textit{Universidad Industrial de Santander}\\
Bucaramanga, Colombia \\
emmanuel2162134@correo.uis.edu.co}
\and
\IEEEauthorblockN{2\textsuperscript{nd} Edwin Vargas}
\IEEEauthorblockA{\textit{Department of Computer Science} \\
\textit{Universidad Industrial de Santander}\\
Bucaramanga, Colombia \\
edwin.vargas4@correo.uis.edu.co}
\and
\IEEEauthorblockN{3\textsuperscript{rd} Henry Arguello}
\IEEEauthorblockA{\textit{Department of Computer Science} \\
\textit{Universidad Industrial de Santander}\\
Bucaramanga, Colombia \\
henarfu@uis.edu.co}
\thanks{This work was supported by the project Sistema general de regal\'ias-Colombia BPIN 2020000100415 with UIS code 8933.}}

\maketitle

\begin{abstract}
The abundant spatial and angular information from light fields has allowed the development of multiple disparity estimation approaches. However, the acquisition of light fields requires high storage and processing cost, limiting the use of this technology in practical applications. To overcome these drawbacks, the compressive sensing (CS) theory has allowed the development of optical architectures to acquire a single coded light field measurement. This measurement is decoded using an optimization algorithm or deep neural network that requires high computational costs. The traditional approach for disparity estimation from compressed light fields requires first recovering the entire light field and then a post-processing step, thus requiring long times. In contrast, this work proposes a fast disparity estimation from a single compressed measurement by omitting the recovery step required in traditional approaches. Specifically, we propose to jointly optimize an optical architecture for acquiring a single coded light field snapshot and a convolutional neural network (CNN) for estimating the disparity maps. Experimentally, the proposed method estimates disparity maps comparable with those obtained from light fields reconstructed using deep learning approaches. Furthermore, the proposed method is $20$ times faster in training and inference than the best method that estimates the disparity from reconstructed light fields.

\end{abstract}


\section{Introduction}

Disparity estimation has been well studied in recent years, since it allows to infer the scene geometry from 2D images.
Some of its most relevant applications include computational 3D modeling of real-life scenarios, face recognition, robotics and medicine \cite{huang20193d, mildenhall2020nerf, chen2017face, de2011human, wu2020robotic}. 
Disparity estimation can be achieved taking advantage of the angular information of the light fields \cite{adelson1991plenoptic, levoy1996light, gortler1996lumigraph, chen2003light, wu2017light}. Light fields collect the amount of light coming from all directions at each spatial point of a physical scene, nevertheless, the huge captured amount of spatial and angular information produces large computational storage and processing costs. Additionally, the multiplexing in microlenses of some of the optical systems also imposes a trade-off between spatial and angular resolution \cite{wu2017light}. To overcome these limitations, different optical architectures \cite{marwah2013compressive, inagaki2018learning, hajisharif2020single} that leverage the CS theory \cite{candes2008introduction} have been proposed for efficient acquisition. Specifically, CS exploits the fact that if the light fields are sparse in some representation basis, they can be compressed \cite{candes2008introduction}.

Once acquired the compressed measurements, the recovery of the light field must be carried out in a digital processing step. Light field reconstruction can be performed using traditional CS algorithms \cite{marwah2013compressive, inagaki2018learning, hajisharif2020single}, or more recently, using deep learning approaches \cite{wu2017light, vadathya2019unified}. Most advanced methods for light field reconstruction proposed to design coded masks in an end-to-end (E2E) approach \cite{guo2020deep, vargas2021time, le2021deep, tateishi2021factorized}. This allows the coded aperture be directly adapted from the training data domain and achieves better performance than traditional approaches. However, the training stage requires long processing times and entails high computational costs. Furthermore, to achieve high quality reconstructions, multiple compressed light field captures are often required.

Traditionally, disparity estimation from compressed measurements requires first an algorithm to recover the full light field and then a  disparity estimation algorithm to obtain the desired disparity. This two-step methodology degrades the quality of the final estimated disparity maps since the reconstruction errors in the first step are propagated to the second step. Furthermore, each step takes long processing times, and performing them sequentially increase the overall time for disparity estimation. Although, some approaches estimate disparity maps in an unsupervised manner as an implicit step in the reconstruction process of light fields from compressed random measurements \cite{vadathya2017learning, vadathya2019unified}, the performance is lower than a supervised method and still requires long processing times also related to their multi-step methodology.

In this work, we propose a fast algorithm to directly estimate the disparity maps from  compressed light fields based on an E2E approach, omitting the reconstruction process. Specifically, we jointly optimize the attenuation values of the coded mask in the optical architecture proposed by \cite{marwah2013compressive} 
and the parameters of a CNN model that decode the sensor measurement to directly obtain the disparity estimation from the compressed measurements.
In addition, we experimentally demonstrate that the proposed method is quantitative and qualitatively comparable to the state-of-the-art methods that use high-quality reconstructed light fields based on deep learning methods. Besides, our method is $20$ times faster in both training and inference stages than the best comparison method based on deep learning as demonstrated in subsection \ref{subsection:results}.

\section{Light Field Compressed Acquisition}
\label{adqusition_model}


To acquire the compressed light field, we consider the optical architecture developed by \cite{marwah2013compressive}. The optical configuration consists of a camera with an objective lens at a distance $d_a$ from the sensor and an attenuation coded mask (ACM) located at a short distance $d_m$ in front of the sensor. Specifically, a light field $F(x, y, u, v)$ is modulated by an ACM $\phi(x, y)$, where $(x, y)$ corresponds to spatial dimension and $(u, v)$ corresponds to the angular dimension.The acquired projections after the light field passes through the ACM can be modeled as

\begin{equation}
	I(x, y) = \iint \phi(x + \tau (u - x), y + \tau (v - y)) F(x, y, u, v) \,du\,dv,
	\label{eq:continuous_lf}
\end{equation}
where $\tau = d_m / d_a$ is the shear of the ACM pattern associated to the input of the scene. Eq. (\ref{eq:continuous_lf}) can be expressed in vector form as

\begin{equation}
	\hat{\mathbf{I}} = \mtx{H}_{\Phi} \mtx{f} + \epsilon,
	\label{eq:simple_lf}
\end{equation}
where $\hat{\mtx{I}} \in \mathbb{R}^m$ is the compressed measurement, $\mtx{f} \in \mathbb{R}^{n}$ is the vector form of the light field, $\mtx{H}_\Phi \in \mathbb{R}^{m \times n} = \rvect{\mtx{H}^{(1)}_\Phi, & \mtx{H}^{(2)}_\Phi, & \cdots, & \mtx{H}^{(S)}_\Phi}$ represents the sensing matrix of the compressive light field photography architecture \cite{marwah2013compressive}, where $\mtx{H}^{(s)}_\Phi \in \mathbb{R}^{m \times m}$ is a diagonal matrix that represents the modulation for the $s$-th angular view, and its non-zero values depend on the discrete representation of the ACM $\Phi \in \mathbb{R}^{m}$. Finally, the compressed measurement of the entire scene can also be viewed as a weighted sum of each modulation of the angular views of the scene as

\begin{equation}
	\hat{\mtx{I}} = \sum_{s = 1}^S \mtx{H}^{(s)}_\Phi \mtx{f}^{(s)} + \epsilon.
	\label{eq:compressed_lf}
\end{equation}
where each $\mtx{f}^{(s)}$ contains the $s$-th angular view and $S$ is the total number of angular views.

\begin{figure}[b!]
    \centering
	\includegraphics[width=.45\textwidth]{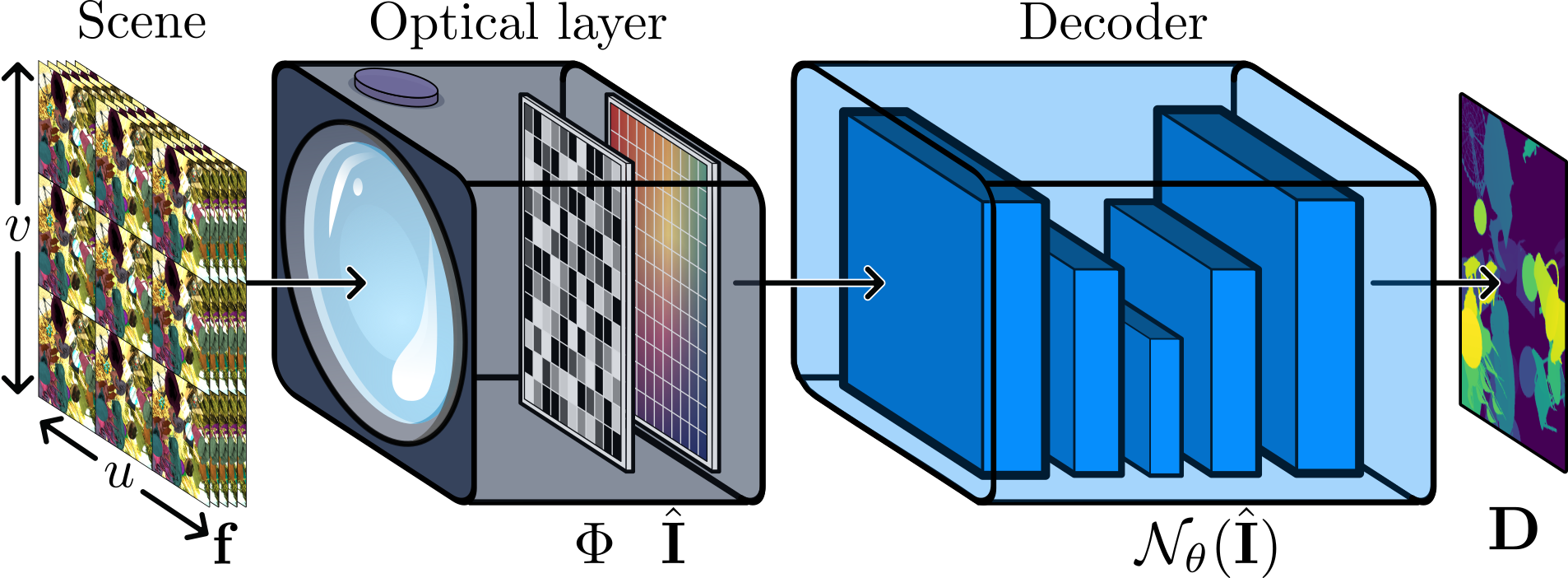}
	\caption{Proposed E2E method. It is divided into two layers: The optical layer generates the compressed measurements from the spatial and angular modulation of the scenes through the ACM. The decoder is focused on estimating the disparity maps from the compressed measurements.}
	\label{figure:e2e}
\end{figure}

\section{Disparity Estimation from Compressed Light Field}

The main idea of the proposed method is to jointly optimize the optical system represented by the ACM $\boldsymbol{\Phi}$ and a electronic decoder represented by a CNN for a fast disparity estimation. The proposed method is divided in two stages: a training stage that leverages a dataset of light fields and corresponding disparity maps to  optimize the ACM and the CNN in a E2E manner; and an inference stage where compressed measurements of testing light fields acquired by the optimized optical system are fed to the trained CNN to estimate the disparity maps (Fig. \ref{figure:e2e}).

\subsection{Training Stage}

For the E2E optimization of the optical encoding and electronic decoder, consider a dataset of light fields $\{\mtx{f}_i\}_{i = 1}^M$ that are modulated by the ACM to generate compressed measurements $\hat{\mtx{I}}_i = \mtx{H}_\Phi \mtx{f}_i$ according to \eqref{eq:compressed_lf}. These measurements are fed to a CNN $\mathbfcal{N}_\theta$ with parameters $\theta$ to generate  estimated disparity maps $\hat{\mtx{D}}_i = \mathbfcal{N}_\theta (\mtx{H}_\Phi \mtx{f}_i)$. Then, to learn the ACM $\Phi$ and the CNN parameters $\theta$ we optimize a cost function $\mathcal{L}$ to minimize the error between the estimated $\hat{\mtx{D}}_i$ and ground truth ${\mtx{D}}_i$ disparity maps

\begin{equation} \label{eq:trainmask}
	\begin{aligned}
		& \{\Phi^\ast, \theta^\ast \} = && \underset{\Phi, \theta}{\text{argmin}} && \frac{1}{M} \sum_{i = 1}^M \mathcal{L} (\mathbfcal{N}_\theta (\mtx{H}_\Phi \mtx{f}_i), \mtx{D}_i) \\
		& && \text{subject to} 	 && \Phi_i \in [0, 1], \hspace{5mm} 1 \le i \le m. \\
	\end{aligned}
\end{equation}
where $\Phi^\ast$ is the optimal ACM and $\theta^\ast$ are the optimal parameters of $\mathcal{N}_\theta$. It is important to highlight that when using a CNN to estimate the disparity, the error given by the loss function $\mathcal{L}$ can be propagated back to the optical system, allowing a joint optimization of the entire model. 


\begin{figure}[t!]
	\includegraphics[width=.9\linewidth]{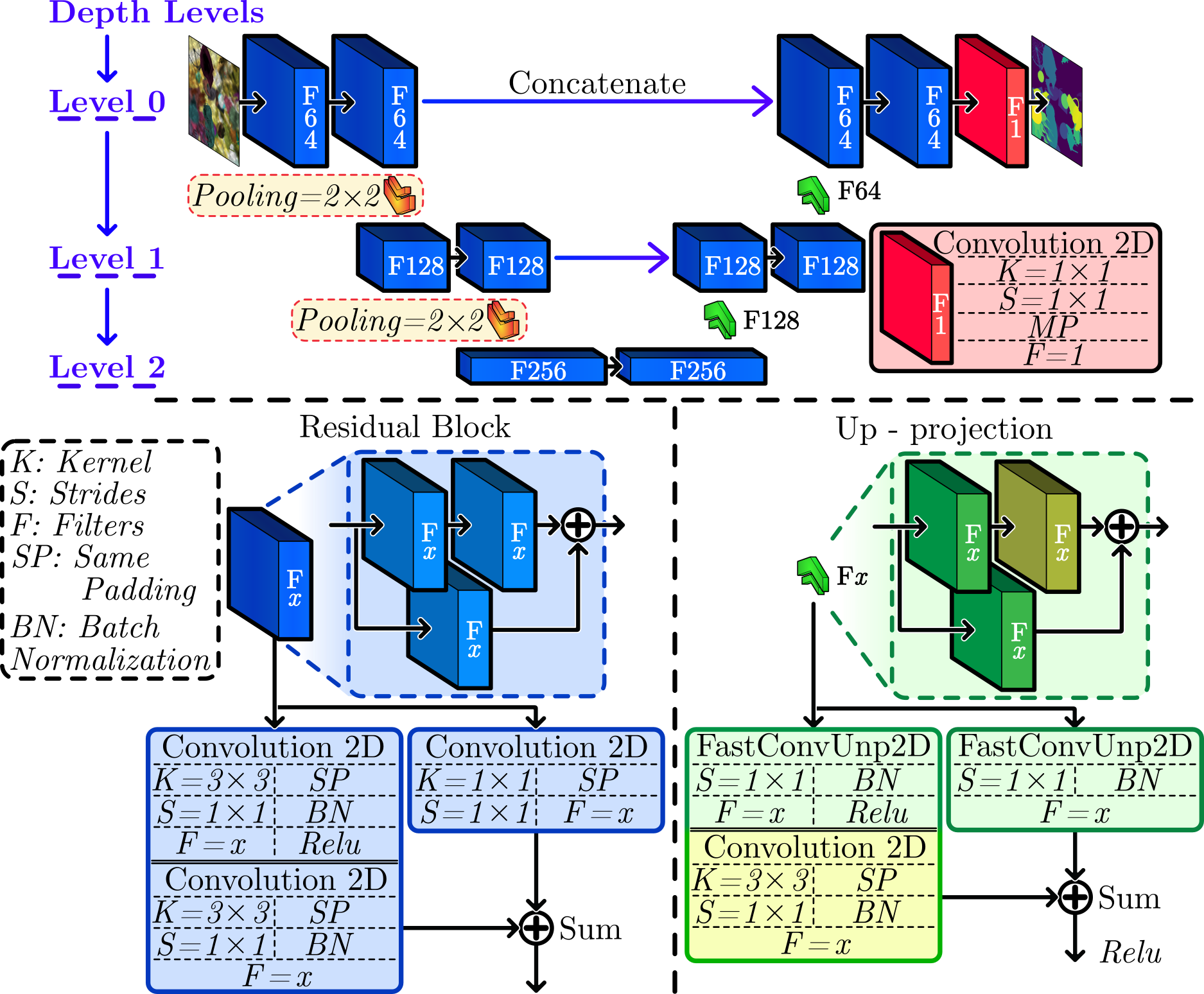}
	\centering
	\caption{CNN model for the decoder of the proposed method. Based on wide U-net architecture where each layer of the network consists of residual blocks, MaxPooling, and up-projections.}
	\label{figure:resupnet}
\end{figure}

\subsection{Inference Stage}

Once the training stage is carried out, the optimized optical parameters are employed to simulate compressed measurements of a light field sampled from a given test set $
	\hat{\mtx{I}}^\ast = \mtx{H}_{\Phi^\ast} \mtx{f},
$
where $\Phi^\ast$ represents the optimized ACM. Then the estimated disparity map $\hat{\mtx{D}}$ is computed from the trained CNN
	$\mtx{\hat{D}} = \mathbfcal{N}_{\theta^\ast} (\hat{\mtx{I}}^\ast ) = \mathbfcal{N}_{\theta^\ast} (\mtx{H}_{\Phi^\ast} \mtx{f})$.
We emphasize here that the trained optical parameter can be translated to a physical device and employ the trained CNN to infer disparity maps from real captures \cite{sitzmann2018end, vargas2021time}.

\section{Simulation Results}


\subsection{Dataset}

The dataset employed in this work was built in \cite{schambach2020multispectral}. It consists of 500 synthetic scenes of RGB light fields with their respective central disparity map. The dataset is divided into 3 parts: 400 samples for training, 50 samples for validation, and 50 samples for testing. To reduce the computational cost, all samples were angularly and spatially reduced to $7 \times 7$ and $256 \times 256$, respectively. Each sample was cropped into patches with the same angular resolution and spatial resolution of $32 \times 32$, for a total of $ 32,000$ patches for the entire dataset.

\begin{figure}[b!]
    \centering
	\includegraphics[width=.4\textwidth]{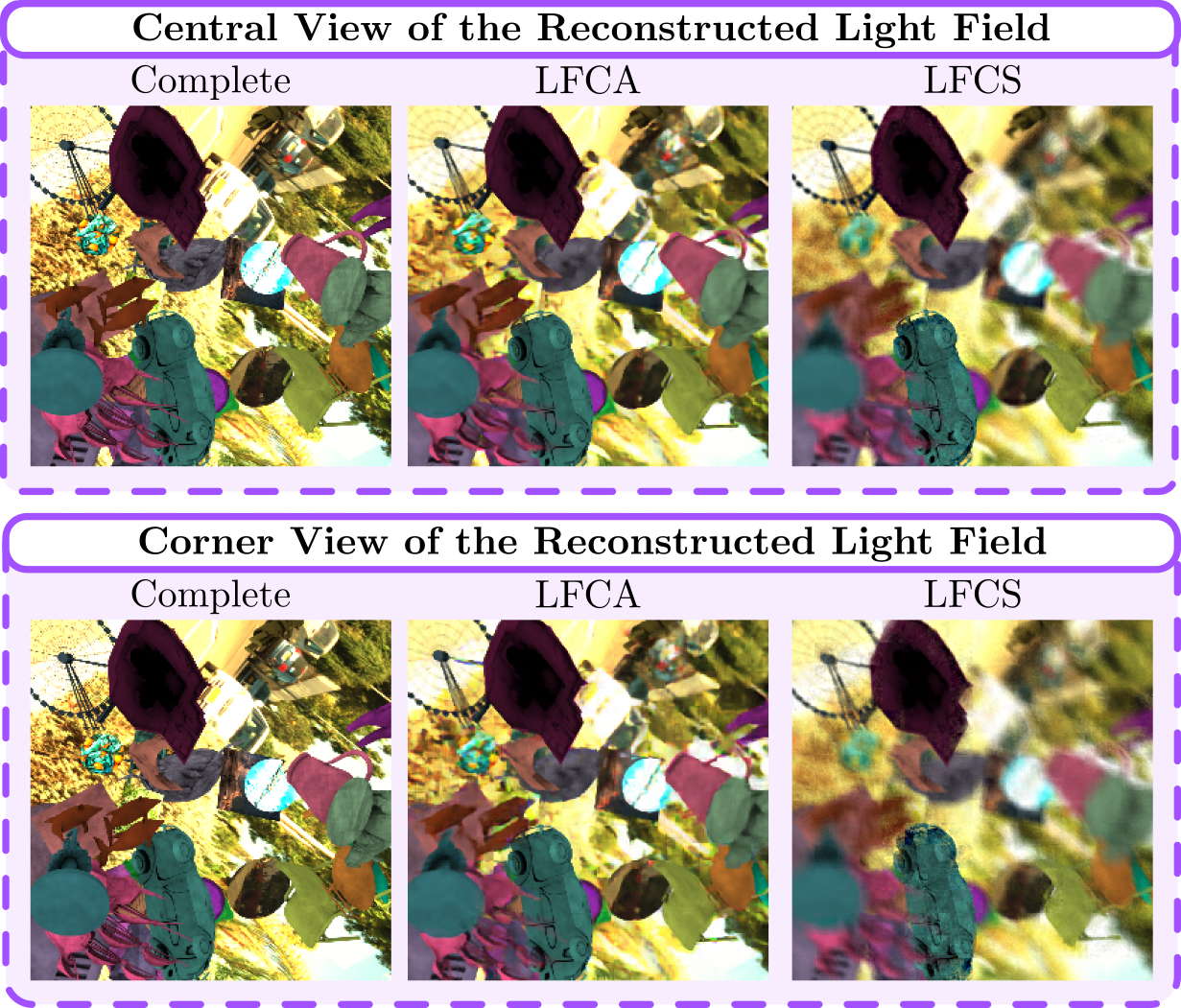}
	\caption{Light field reconstructions. Center view sample and view sample located in a corner of the light field with the respective LFCS and LFCA reconstructions where complete is the synthetic sample.}	
	\label{figure:rec_lf}
\end{figure}

\subsection{Decoder}
Inspired by the CNNs \cite{laina2016deeper, harsanyi2018hybrid} based on the U-net architecture \cite{ronneberger2015u} with residual blocks \cite{he2016deep} and ascending projections \cite{laina2016deeper}, we propose the CNN $\mathcal{N}_\theta$ shown in Fig. \ref{figure:resupnet} to estimate the disparity maps from the optimized projections. Furthermore, considering that deep CNNs can suffer the feature reuse problem \cite{zagoruyko2016wide}, the proposed CNN is a shallow network with $3$ layers and we increase the number of filters. We experimentally find that the proposed network works better for the decoding of light fields than deeper CNNs.
\subsection{Loss Function and Metrics}

To quantitatively evaluate the results obtained, various metrics were used to accurately quantify the pixel-to-pixel differences between the true disparity maps and the predictions. To train and evaluate the proposed method, we employ as loss function $\mathcal{L}$ in \eqref{eq:trainmask} the pseudo-Huber loss, allowing to control the outliers between the true disparity maps and their estimations \cite{gokcesu2021generalized}. The other metrics used to evaluate the quality of the predictions are the mean square error (MSE) and mean absolute error (MAE), total variation (TV) and the bad pixel ratio (BadPix) \cite{honauer2016dataset}.

\subsection{Experimental setup}

For the proposed method, we consider two configurations: an E2E configuration where the ACM and the digital decoder weights are jointly learned, and a CNN configuration where the ACM is fixed and random. The random ACM was elaborated using a normal distribution function. It is worth noting that the CNN requires as input the single coded snapshot measurement of the light fields, thus the required processing would be $7 \times 7$ times smaller compared to the processing of the full light fields with $7 \times 7$ angular resolution and $256 \times 256$ spatial resolution.
An augmentation was applied to both configurations with the dataset during the training, which consist of random horizontal flip, rotation and gamma stretch applied to the channels of the light fields. For the training of the proposed method (both configurations), $100$ epochs were performed, with mini-batches of $16$ samples. The root mean square propagation (RMSprop) optimizer with a fixed learning rate of $5e{-4}$ was used to solve the optimization problem in  \eqref{eq:trainmask}.

\begin{figure}[t!]
	\includegraphics[width=.3\textwidth]{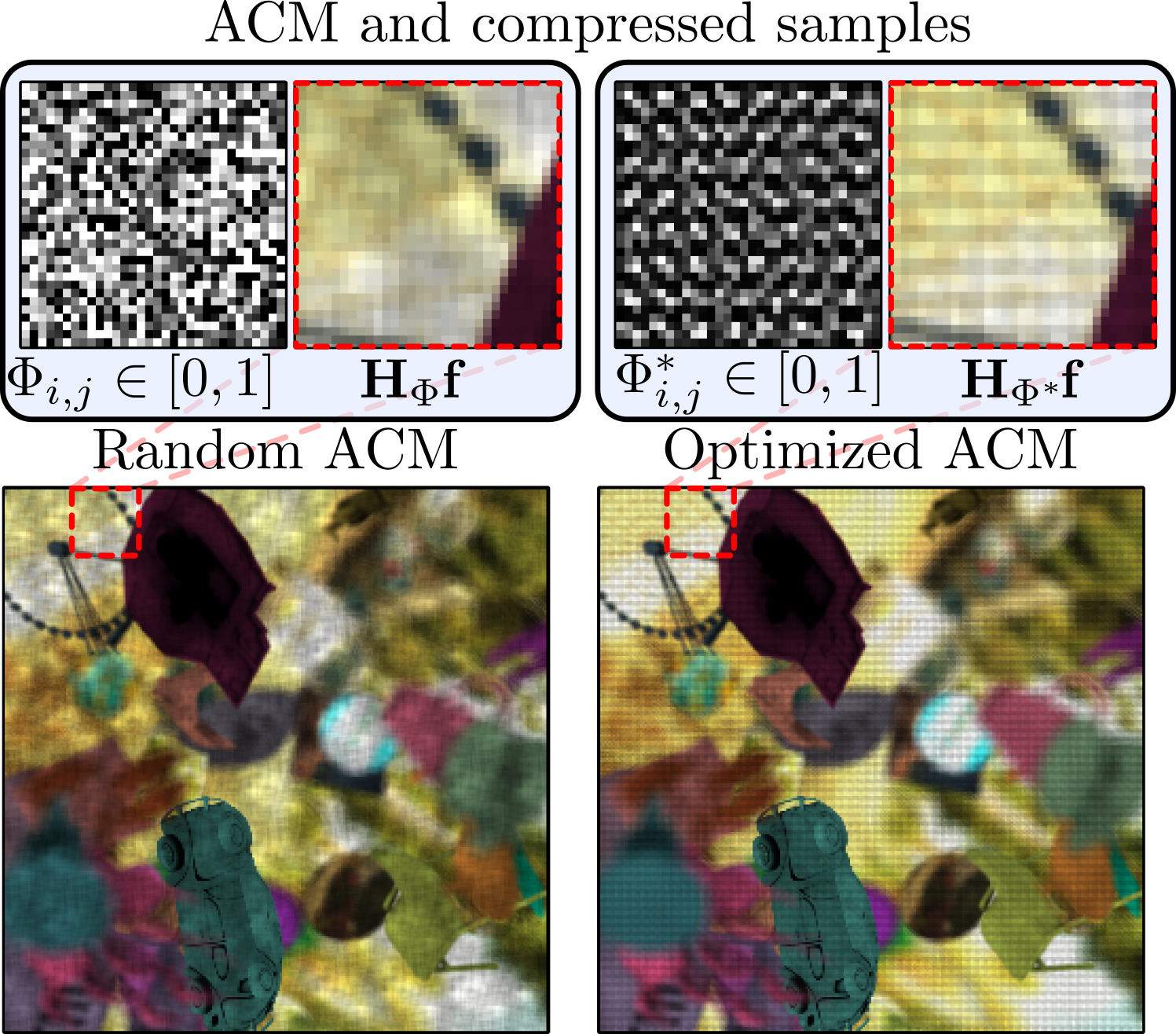}
	\centering
	\caption{ACMs and corresponding compressed measurements. Left: Random ACM employed to run the CNN configuration. Right: E2E optimized ACM.}
	\label{figure:masks}
\end{figure}

\begin{figure*}[t!]
	\includegraphics[width=0.85\textwidth]{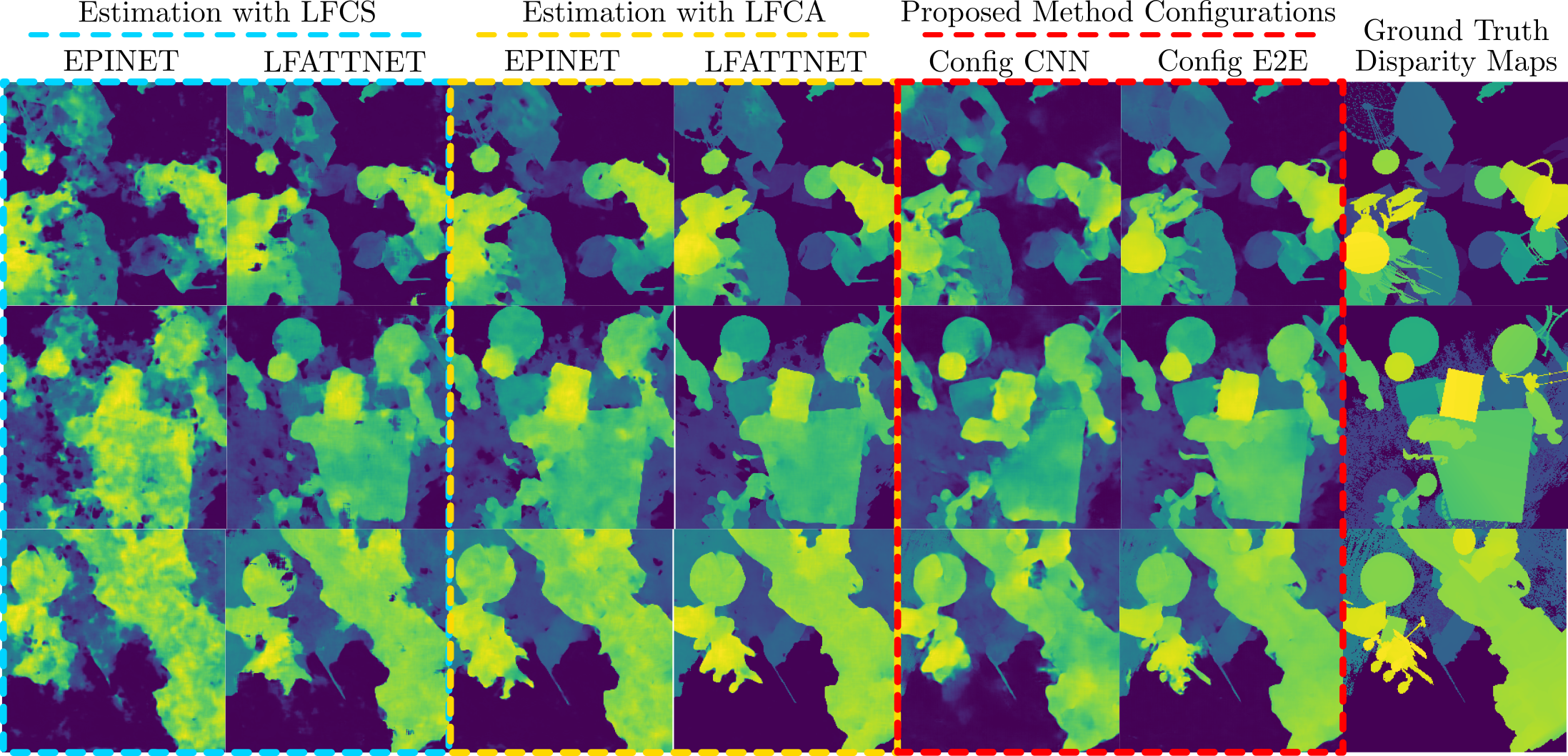}
	\centering
	\caption{Qualitative results of the estimation of disparity maps. Dashed cyan and yellow blocks are the estimations by comparison models, and the dashed red block are the estimations by proposed method.}
	\label{figure:full_disp}
\end{figure*}

\begin{table*}[t!]
    \centering
    \caption{Quantitative results of the estimation of disparity maps.}
	\scalebox{1.15}{
		\begin{tabular}{ccccccccc}
			\hline
			&   &     &     &    &    &    & \multicolumn{2}{c}{Time}     \\ \cline{8-9} 
			\multirow{-2}{*}{\begin{tabular}[c]{@{}c@{}}Configurations and \\ Comparison Models\end{tabular}} & \multirow{-2}{*}{\begin{tabular}[c]{@{}c@{}}Pseudo\\ Huber\end{tabular}} & \multirow{-2}{*}{MAE} & \multirow{-2}{*}{MSE} & \multirow{-2}{*}{BadPix01} & \multirow{-2}{*}{BadPix03} & \multirow{-2}{*}{BadPix07} & \begin{tabular}[c]{@{}c@{}}Training\\ (hours)\end{tabular} & \begin{tabular}[c]{@{}c@{}}Inference\\ (minutes)\end{tabular} \\ \hline
			\rowcolor[HTML]{E0E0E0} 
			Reference & 0.00035  & 0.00488  & 0.00055  & 4.6237   & 3.0980   & 2.0553   & 18.5   & 1.1  \\ \hline
			Epinet + LFCS   & 0.00410 & 0.04541   & 0.00680   & 61.1211  & 44.5674  & 23.3791  & \textit{20.944}   & 30.16    \\
			Epinet + LFCA   & \textit{0.00172}    & \textit{0.02240}& \textit{0.00291}& \textit{41.0658}     & \textit{16.5027}     & \textit{7.3672}& 104.457     & \underline{6.78}  \\ \hline
			Lfattnet + LFCS & 0.00333 & 0.03632   & 0.00580   & 50.3294  & 27.7080  & 14.6844  & 26.638 & 30.45    \\
			Lfattnet + LFCA & \underline{0.00134} & \textbf{0.01729}& \underline{0.00229}   & \textbf{29.4294}     & \textbf{11.0752}     & \underline{5.8249}   & 110.512  & \textit{7.61}  \\ \hline
			Configuration CNN    & 0.00367 & 0.04580   & 0.00571   & 73.8832  & 43.4794  & 19.1005  & \textbf{2.639} & \multirow{2}{*}{\textbf{0.33}}   \\
			Configuration E2E    & \textbf{0.00127}    & \underline{0.01891}   & \textbf{0.00204}& \underline{38.3157}  & \underline{13.9681}  & \textbf{5.6851}& \underline{5.416} &      \\ \hline
		\end{tabular}
	}
	\label{table:disp_result_table}
\end{table*}

We compare the proposed method with the strategy that reconstructs the light field and then estimates the disparity map. Two different types of light field reconstructions were considered: a reconstruction based on traditional iterative algorithm (LFCS) \cite{marwah2013compressive} and a reconstruction based on deep learning (LFCA) \cite{guo2020deep}, as shown in Fig. \ref{figure:rec_lf}. We note that the LFCA method employs a different acquisition system and we adapt this method to employ the architecture presented in section \ref{adqusition_model}. For disparity estimation, we employ  two state-of-the-art CNN models, Epinet \cite{shin2018epinet} and  Lfattnet \cite{tsai2020attention}. We train these networks from scratch using the reconstructed light fields.
The experiments and comparison models were trained with Tensorflow \cite{abadi2016tensorflow} using an NVIDIA GeForce RTX 3090 video card. The LFCS method was trained in MATLAB without GPU usage.

\subsection{Simulation Results}
\label{subsection:results}

The ACMs of the proposed two configurations are shown in Fig. \ref{figure:masks}. Following the same practice in \cite{marwah2013compressive}, the ACM has the same image size but with a periodic structure built with a mask of $32 \times 32$. It is observed that the optimized ACM learns a semi-uniform grid pattern that results in compressed measurements whit less distortion than the measurements acquired with the random ACM. In this way, the proposed CNN can extract richer features for the disparity estimation task from the designed compressed light field measurements than the non-designed ones.

The quantitative evaluation and qualitative performance of the estimated disparity maps with the proposed method and the comparison methods are reported in the table \ref{table:disp_result_table} and in the Fig. \ref{figure:full_disp}, respectively. The table results correspond to the average of the best performance. The ``Reference'' in the table refers to the highest performance of the \underline{\textbf{Lfattnet model}} trained using the full light field samples. Therefore, it will not be taken into account for comparison. Table \ref{table:disp_result_table} shows that the performance of the proposed method using the E2E configuration is comparable to the Lfattnet model having as input the reconstructed light field from LFCA. The lower performance was achieved with the approaches that employ LFCS due to  the low quality of its reconstructions.

The most notable gain of the proposed method is found in the training time, where the E2E configuration takes $195$ seconds by epoch. Therefore, for $100$ epochs it takes a total time of $5.416$ hours. Lfattnet trained with reconstructions from LFCA takes a total of $18.5$ hours being $3.4$ times slower than the proposed E2E configuration. Furthermore, considering that the light field reconstruction using LFCA took approximately $4$ days, the proposed method is $20$ times faster in training than the best competitive result obtained using Lfattnet and LFCA. Applying the same analysis to Epinet, the E2E configuration is $2.3$ times faster and taking into account the reconstruction using LFCA, it would be $19$ times faster. It can be seen that training with the CNN configuration is the fastest to be performed as expected since it does not require the optimization of the ACM. When analyzing the inference times, it is observed that both configurations of the proposed method are faster in this process than all the comparison methods. Specifically, the E2E configuration manages to be $23$ times faster than the best comparison method.

\section{Conclusions}

A fast method was proposed to estimate disparity from a single compressed measurement of light fields using an E2E approach. The optical coding of the light field photography system is optimized jointly with a CNN for direct disparity estimation from the compressed measurement.
The results obtained suggest that the proposed E2E method is comparable with state-of-the-art models for disparity estimation that employ as input reconstructed light fields by compressive sensing or deep learning algorithms. Finally, the proposed method with E2E configuration proved to be $20$ times faster in training and inference than the best comparison method that separately learns two deep networks for recovering the light fields and estimating the disparity maps.

\bibliographystyle{IEEEbib}
\bibliography{biblio}

\begin{thebibliography}{10}
\providecommand{\url}[1]{#1}
\csname url@samestyle\endcsname
\providecommand{\newblock}{\relax}
\providecommand{\bibinfo}[2]{#2}
\providecommand{\BIBentrySTDinterwordspacing}{\spaceskip=0pt\relax}
\providecommand{\BIBentryALTinterwordstretchfactor}{4}
\providecommand{\BIBentryALTinterwordspacing}{\spaceskip=\fontdimen2\font plus
\BIBentryALTinterwordstretchfactor\fontdimen3\font minus
  \fontdimen4\font\relax}
\providecommand{\BIBforeignlanguage}[2]{{%
\expandafter\ifx\csname l@#1\endcsname\relax
\typeout{** WARNING: IEEEtran.bst: No hyphenation pattern has been}%
\typeout{** loaded for the language `#1'. Using the pattern for}%
\typeout{** the default language instead.}%
\else
\language=\csname l@#1\endcsname
\fi
#2}}
\providecommand{\BIBdecl}{\relax}
\BIBdecl

\bibitem{huang20193d}
H.~Huang, A.~Kuhn, M.~Michelini, M.~Schmitz, and H.~Mayer, ``3d urban scene
  reconstruction and interpretation from multisensor imagery,'' in
  \emph{Multimodal Scene Understanding}.\hskip 1em plus 0.5em minus 0.4em\relax
  Elsevier, 2019, pp. 307--340.

\bibitem{mildenhall2020nerf}
B.~Mildenhall, P.~P. Srinivasan, M.~Tancik, J.~T. Barron, R.~Ramamoorthi, and
  R.~Ng, ``Nerf: Representing scenes as neural radiance fields for view
  synthesis,'' in \emph{European conference on computer vision}.\hskip 1em plus
  0.5em minus 0.4em\relax Springer, 2020, pp. 405--421.

\bibitem{chen2017face}
C.-H. Chen and R.~Chellappa, ``Face recognition using an outdoor camera
  network,'' in \emph{Human Recognition in Unconstrained Environments}.\hskip
  1em plus 0.5em minus 0.4em\relax Elsevier, 2017, pp. 31--54.

\bibitem{de2011human}
G.~De~Cubber and D.~Doroftei, ``Human victim detection and stereo-based terrain
  traversability analysis for behaviour-based robot navigation,'' in
  \emph{Using Robots in Hazardous Environments}.\hskip 1em plus 0.5em minus
  0.4em\relax Elsevier, 2011, pp. 476--498.

\bibitem{wu2020robotic}
L.~Wu, A.~Jaiprakash, A.~K. Pandey, D.~Fontanarosa, Y.~Jonmohamadi, M.~Antico,
  M.~Strydom, A.~Razjigaev, F.~Sasazawa, J.~Roberts \emph{et~al.}, ``Robotic
  and image-guided knee arthroscopy,'' in \emph{Handbook of Robotic and
  Image-Guided Surgery}.\hskip 1em plus 0.5em minus 0.4em\relax Elsevier, 2020,
  pp. 493--514.

\bibitem{adelson1991plenoptic}
E.~H. Adelson, J.~R. Bergen \emph{et~al.}, \emph{The plenoptic function and the
  elements of early vision}.\hskip 1em plus 0.5em minus 0.4em\relax Vision and
  Modeling Group, Media Laboratory, Massachusetts Institute of~…, 1991,
  vol.~2.

\bibitem{levoy1996light}
M.~Levoy and P.~Hanrahan, ``Light field rendering,'' in \emph{Proceedings of
  the 23rd annual conference on Computer graphics and interactive techniques},
  1996, pp. 31--42.

\bibitem{gortler1996lumigraph}
S.~J. Gortler, R.~Grzeszczuk, R.~Szeliski, and M.~F. Cohen, ``The lumigraph,''
  in \emph{Proceedings of the 23rd annual conference on Computer graphics and
  interactive techniques}, 1996, pp. 43--54.

\bibitem{chen2003light}
W.-C. Chen, ``Light field mapping: Efficient representation of surface light
  fields,'' \emph{Energy, simulation-training, ocean engineering, and
  instrumentation: research papers of the Link Foundation fellows}, p.~89,
  2003.

\bibitem{wu2017light}
G.~Wu, B.~Masia, A.~Jarabo, Y.~Zhang, L.~Wang, Q.~Dai, T.~Chai, and Y.~Liu,
  ``Light field image processing: An overview,'' \emph{IEEE J. Sel. Topics
  Signal Process.}, vol.~11, no.~7, pp. 926--954, 2017.

\bibitem{marwah2013compressive}
K.~Marwah, G.~Wetzstein, Y.~Bando, and R.~Raskar, ``Compressive light field
  photography using overcomplete dictionaries and optimized projections,''
  \emph{ACM Transactions on Graphics (TOG)}, vol.~32, no.~4, pp. 1--12, 2013.

\bibitem{inagaki2018learning}
Y.~Inagaki, Y.~Kobayashi, K.~Takahashi, T.~Fujii, and H.~Nagahara, ``Learning
  to capture light fields through a coded aperture camera,'' in
  \emph{Proceedings of the European Conference on Computer Vision (ECCV)},
  2018, pp. 418--434.

\bibitem{hajisharif2020single}
S.~Hajisharif, E.~Miandji, C.~Guillemot, and J.~Unger, ``Single sensor
  compressive light field video camera,'' in \emph{Computer Graphics Forum},
  vol.~39, no.~2.\hskip 1em plus 0.5em minus 0.4em\relax Wiley Online Library,
  2020, pp. 463--474.

\bibitem{candes2008introduction}
E.~J. Cand{\`e}s and M.~B. Wakin, ``An introduction to compressive sampling,''
  \emph{IEEE Signal Process. Mag.}, vol.~25, no.~2, pp. 21--30, 2008.

\bibitem{vadathya2019unified}
A.~K. Vadathya, S.~Girish, and K.~Mitra, ``A unified learning-based framework
  for light field reconstruction from coded projections,'' \emph{IEEE Trans.
  Comput. Imag.}, vol.~6, pp. 304--316, 2019.

\bibitem{guo2020deep}
M.~Guo, J.~Hou, J.~Jin, J.~Chen, and L.-P. Chau, ``Deep spatial-angular
  regularization for compressive light field reconstruction over coded
  apertures,'' in \emph{European Conference on Computer Vision}.\hskip 1em plus
  0.5em minus 0.4em\relax Springer, 2020, pp. 278--294.

\bibitem{vargas2021time}
E.~Vargas, J.~N. Martel, G.~Wetzstein, and H.~Arguello, ``Time-multiplexed
  coded aperture imaging: Learned coded aperture and pixel exposures for
  compressive imaging systems,'' in \emph{Proceedings of the IEEE/CVF
  International Conference on Computer Vision}, 2021, pp. 2692--2702.

\bibitem{le2021deep}
G.~Le~Guludec, E.~Miandji, and C.~Guillemot, ``Deep light field acquisition
  using learned coded mask distributions for color filter array sensors,''
  \emph{IEEE Trans. Comput. Imag.}, vol.~7, pp. 475--488, 2021.

\bibitem{tateishi2021factorized}
K.~Tateishi, K.~Sakai, C.~Tsutake, K.~Takahashi, and T.~Fujii, ``Factorized
  modulation for singleshot lightfield acquisition,'' in \emph{2021 IEEE
  International Conference on Image Processing (ICIP)}.\hskip 1em plus 0.5em
  minus 0.4em\relax IEEE, 2021, pp. 3253--3257.

\bibitem{vadathya2017learning}
A.~K. Vadathya, S.~Cholleti, G.~Ramajayam, V.~Kanchana, and K.~Mitra,
  ``Learning light field reconstruction from a single coded image,'' in
  \emph{2017 4th IAPR Asian Conference on Pattern Recognition (ACPR)}.\hskip
  1em plus 0.5em minus 0.4em\relax IEEE, 2017, pp. 328--333.

\bibitem{sitzmann2018end}
V.~Sitzmann, S.~Diamond, Y.~Peng, X.~Dun, S.~Boyd, W.~Heidrich, F.~Heide, and
  G.~Wetzstein, ``End-to-end optimization of optics and image processing for
  achromatic extended depth of field and super-resolution imaging,'' \emph{ACM
  Transactions on Graphics (TOG)}, vol.~37, no.~4, pp. 1--13, 2018.

\bibitem{schambach2020multispectral}
M.~Schambach and M.~Heizmann, ``A multispectral light field dataset and
  framework for light field deep learning,'' \emph{IEEE access}, vol.~8, pp.
  193\,492--193\,502, 2020.

\bibitem{laina2016deeper}
I.~Laina, C.~Rupprecht, V.~Belagiannis, F.~Tombari, and N.~Navab, ``Deeper
  depth prediction with fully convolutional residual networks,'' in \emph{2016
  Fourth international conference on 3D vision (3DV)}.\hskip 1em plus 0.5em
  minus 0.4em\relax IEEE, 2016, pp. 239--248.

\bibitem{harsanyi2018hybrid}
K.~Hars{\'a}nyi, A.~Kiss, A.~Majdik, and T.~Szir{\'a}nyi, ``A hybrid cnn
  approach for single image depth estimation: A case study,'' in
  \emph{International Conference on Multimedia and Network Information
  System}.\hskip 1em plus 0.5em minus 0.4em\relax Springer, 2018, pp. 372--381.

\bibitem{ronneberger2015u}
O.~Ronneberger, P.~Fischer, and T.~Brox, ``U-net: Convolutional networks for
  biomedical image segmentation,'' in \emph{International Conference on Medical
  image computing and computer-assisted intervention}.\hskip 1em plus 0.5em
  minus 0.4em\relax Springer, 2015, pp. 234--241.

\bibitem{he2016deep}
K.~He, X.~Zhang, S.~Ren, and J.~Sun, ``Deep residual learning for image
  recognition,'' in \emph{Proceedings of the IEEE conference on computer vision
  and pattern recognition}, 2016, pp. 770--778.

\bibitem{zagoruyko2016wide}
S.~Zagoruyko and N.~Komodakis, ``Wide residual networks,'' \emph{arXiv preprint
  arXiv:1605.07146}, 2016.

\bibitem{gokcesu2021generalized}
K.~Gokcesu and H.~Gokcesu, ``Generalized huber loss for robust learning and its
  efficient minimization for a robust statistics,'' \emph{arXiv preprint
  arXiv:2108.12627}, 2021.

\bibitem{honauer2016dataset}
K.~Honauer, O.~Johannsen, D.~Kondermann, and B.~Goldluecke, ``A dataset and
  evaluation methodology for depth estimation on 4d light fields,'' in
  \emph{Asian Conference on Computer Vision}.\hskip 1em plus 0.5em minus
  0.4em\relax Springer, 2016, pp. 19--34.

\bibitem{shin2018epinet}
C.~Shin, H.-G. Jeon, Y.~Yoon, I.~S. Kweon, and S.~J. Kim, ``Epinet: A
  fully-convolutional neural network using epipolar geometry for depth from
  light field images,'' in \emph{Proceedings of the IEEE Conference on Computer
  Vision and Pattern Recognition}, 2018, pp. 4748--4757.

\bibitem{tsai2020attention}
Y.-J. Tsai, Y.-L. Liu, M.~Ouhyoung, and Y.-Y. Chuang, ``Attention-based view
  selection networks for light-field disparity estimation,'' in
  \emph{Proceedings of the AAAI Conference on Artificial Intelligence},
  vol.~34, no.~07, 2020, pp. 12\,095--12\,103.

\bibitem{abadi2016tensorflow}
M.~Abadi, P.~Barham, J.~Chen, Z.~Chen, A.~Davis, J.~Dean, M.~Devin,
  S.~Ghemawat, G.~Irving, M.~Isard \emph{et~al.}, ``Tensorflow: A system for
  large-scale machine learning,'' in \emph{12th $\{$USENIX$\}$ symposium on
  operating systems design and implementation ($\{$OSDI$\}$ 16)}, 2016, pp.
  265--283.

\end{thebibliography}

\end{document}